%% file: main.tex
\documentclass[acmsmall,screen,]{format/acmart}

\usepackage{amsmath}
\usepackage{booktabs}
\usepackage{graphicx}
\usepackage{microtype}
\usepackage{tabularx}
\usepackage{xspace}
\usepackage{xcolor}
\usepackage{enumitem}
\usepackage{tikz}
\usetikzlibrary{arrows.meta,backgrounds,fit,positioning}

\newcommand{\name}{TRUSS\xspace}
\newcommand{\fullname}{\textbf{T}ask-\textbf{R}eliable and \textbf{U}ser-\textbf{S}afe \textbf{S}kill Generation\xspace}

\setcopyright{acmlicensed}
\copyrightyear{2027}
\acmYear{2027}
\acmConference[FSE 2027]{The ACM International Conference on the Foundations
  of Software Engineering}{July 12--16, 2027}{Shenzhen, China}
\acmBooktitle{Proceedings of the ACM International Conference on the Foundations
  of Software Engineering, July 12--16, 2027, Shenzhen, China}
\acmDOI{}
\acmISBN{}

\title{TRUSS: Towards Task-Reliable and User-Safe Automated Agent Skill Generation}

\author{Zhibo Zhang}
\orcid{0009-0008-6447-1756}
\affiliation{%
\institution{Huazhong University of Science and Technology}
\city{Wuhan}
\country{China}
}
\email{zhangzhibom@hust.edu.cn}

\author{Zhen Ouyang}
\orcid{0009-0008-8393-1455}
\affiliation{%
\institution{Huazhong University of Science and Technology}
\city{Wuhan}
\country{China}
}
\email{cookingmaster0920@gmail.com}

\author{Ling Shi}
\orcid{0000-0002-2023-0247}
\affiliation{%
\institution{Nanyang Technological University}
\city{Singapore}
\country{Singapore}
}
\email{lshi.academic@gmail.com}

\author{Kailong Wang}
\authornote{Corresponding Author.}
\orcid{0000-0002-3977-6573}
\affiliation{%
\institution{Huazhong University of Science and Technology}
\city{Wuhan}
\country{China}
}
\email{wangkl@hust.edu.cn}

\begin{document}

\begin{abstract}
Agent Skills package reusable natural language procedures with executable resources, enabling software agents to acquire task specific capabilities without model adaptation. Automatically generating such Skills can improve task performance, yet evaluating a candidate solely from its artifact or final task outcome leaves unresolved which actions the equipped agent will perform and which side effects those actions will produce. We present \name, an evidence guided framework for generating functionally effective and safety reliable Agent Skills. \name first inspects functional claims against source and domain evidence while evaluating the complete artifact under nine predefined safety properties. Candidates admitted by this static gate are loaded by a shadow agent inside a Controllable Execution Environment, where brokered tools expose requested actions to policy enforcement and record their results as provenance preserving execution traces. Functional failures and property violations are linked back to the responsible Skill content and used to guide iterative refinement. 

We evaluate \name on 168 SkillInject artifacts, 155 SkillSafetyBench cases, and all 187 tasks in SkillGenBench. \name achieves 100.00\% precision and recall in vulnerability detection. Repair reduces attack success from 38.71\% to 19.35\% with GPT 5.5 and from 46.45\% to 29.68\% with GPT 5.4, with zero attack regression. For Skill generation, \name raises task effectiveness from 17.11\% without Skills to 52.94\%, while increasing the benchmark Security rate from 50.80\% to 100.00\%. These results show that execution evidence can expose behavioral failures missed by artifact inspection and can guide Skill generation toward jointly verified functional and safety outcomes.

\end{abstract}



\maketitle

\input{sections/introduction}
\input{sections/related_work}

\input{sections/methodology}
\input{sections/evaluation}

\input{sections/conclusion}

\bibliographystyle{format/ACM-Reference-Format}
\bibliography{reference}

\end{document}

%% file: sections/introduction.tex
\section{Introduction}
\label{secintroduction}

Large Language Model (LLM)-based agents are rapidly moving beyond virtual planning toward direct execution for real-world tasks. To better deal with some specialized or process-oriented scenarios, developers extend these agents by packaging reusable workflows, domain-specific knowledge, and tool integrations into file-based instructions, which can effectively improve agent performance without model tuning~\cite{wang2023voyager,nottingham2024skillset,li2026skillsbench}. 
We refer to such reusable packages as \textbf{\textit{skills}}.
Currently, skill authoring relies heavily on manual construction.
Producing a skill entails translating domain procedures into instructions that an agent can interpret and execute reliably across varying tasks and environments, which demands substantial expertise and repeated behavioral validation.
With the sharp increase of the number and diversity of target tasks, automated skill construction has become an important research problem ~\cite{nottingham2024skillset, alzubi2026evoskill, liang2026skillnet, zhou2026skillgenbench}.

\emph{Functional Effectiveness} requires the skill to produce a verified improvement over the base agent on its intended task. Recent works have already adopted paired evaluation as a necessary assessment protocol, which ensures performance improvements on the intended task do not come at the expense of tasks the base agent already completes successfully ~\cite{li2026skillsbench}. 
\emph{Safety Reliability} requires that every action induced by a skill accomplish its intended functionality without introducing security-relevant behavior beyond the intended and user-authorized scope.
Current agent systems grant any installed skills with full local user privileges, with minimal scrutiny or interactive confirmation. Recent security studies have shown that unsafe skills can introduce serious vulnerabilities into LLM-based agents, exposing users and their local environments to substantial security risks \citep{liu2026skillswild}. Major agent providers like OpenAI and Anthropic have explicitly warned that vulnerabilities in third-party or externally sourced skills can lead to harmful actions and potentially severe damage to users' local systems \citep{openai2026codexsecurity, anthropic2025agentskills}.
Reliable generated skills should therefore not expose the local system to potential security risks during execution pipeline, such as executing unauthorized system commands, invoking high-risk tools, accessing malicious or untrusted links, or exfiltrating sensitive data.

However, \emph{Functional Effectiveness} and \emph{Safety Reliability} interact through the operational specificity of a skill. Detailed procedures and executable helpers can increase task success by reducing ambiguity, which at the same time expand the action space available to the agent and introduce uncertain behaviors. Generation pipelines that rely primarily on static analysis of instruction text often struggle to identify the optimal trade-off between functional effectiveness and safety reliability. For example, specialized skills always instruct the agent to retrieve an installation fragment and later executes it. The resulting download and execution chain emerges only after multiple individually plausible actions.
Existing skills auto-generation efforts often conduct static checks on the functionality or security of the generated skills using only a few attributes, and fail to assess their state in dynamic execution scenarios~\cite{nottingham2024skillset, zhou2026skillgenbench, paz2026skillspector}. This results in the generation iteration process not fully capturing the performance of the skills, and the generated skills may expose more problems in actual scenarios. 

In this paper, we present \textbf{\name} (\fullname), an automated skill construction framework that incorporates dynamic execution evidence. Building upon prior work that relies on static property checking, \name goes further by observing candidate skills inside a Controllable Execution Environment (CEE) and using the resulting execution traces to guide iterative refinement. \name supports both the task-conditioned regime, where the target task is available during generation, and the task-agnostic regime, where skills are constructed from source materials without a predetermined downstream task. By abstracting execution traces in the CEE into property-level evidence and combining this evidence with predefined static checks, \name provides functional and safety guarantees for generated skills through a hybrid static-dynamic verification process.
By evaluation across diverse benchmarks and metrics, \name demonstrates consistent skill-generation improvements in function and security performance. 
For vulnerability detection, \name reaches 100\% precision and recall on matched clean/injected skill pairs. Moreover, repairing test on SkillSafetyBench shows \name reduces harmful rate from 38.70\% to 19.35\% (19.35\% elimination) with GPT-5.5, and from 46.45\% to 29.68\% (16.77\% elimination) with GPT-5.4, with no regression on previously successful tasks. Across 187 generation tasks under two input regimes, \name consistently delivers verified improvement over the base agent while maintaining reuse and runtime security.

This paper makes the following contributions.

\begin{itemize}
    \item Our empirical study reveals that static skill screening fails to detect a significant set of runtime vulnerabilities, underscoring the need to complement artifact-level checks with execution-aware verification.

    \item We develop \name, an automated generation and refinement framework, which moves beyond static inspection of the skill artifact to the behavior induced once the artifact is loaded and executed by an agent. 

    \item By comprehensive evaluation across 3 agents and 5 models, \name exhibits significant performance improvements on multiple aspects, indicating that runtime evidence closes detection errors left by static inspection and guides repairs that substantially reduce attack success while preserving previously successful task behavior.
\end{itemize}

%% file: sections/related_work.tex
\section{Related Work}
\label{sec:related_work}
\subsection{Automated Skill Generation}
\label{subsec:related_work:skill_generation}

Automated skill generation seeks to transform task experience and external knowledge into reusable procedural assets that improve an agent without updating its model parameters. 
Execution guided methods evaluate candidate skills through downstream agent performance. Skill Set Optimization constructs transferable skills from trajectories associated with high environment rewards~\citep{nottingham2024skillset}. EvoSkill analyzes failed executions and retains generated skill folders that improve performance on a held out validation set~\citep{alzubi2026evoskill}. These methods expose candidate skills to executable agent environments and use observed task outcomes to guide subsequent generation. Their selection objectives are defined by functional rewards or answer correctness. A candidate can therefore receive a positive refinement signal after producing the expected result even when its trajectory contains unauthorized file access, dangerous command execution, or unintended network communication. Their evidence coverage also follows fixed task partitions and task specific verifiers. Environment conditioned branches and security relevant behaviors outside the exercised scenarios provide no signal to the generation process.

Another direction evaluates multiple properties of the generated artifact before deployment. SkillNet constructs skills from trajectories, repositories, documents, and user instructions, and assesses their safety, completeness, executability, maintainability, and cost~\citep{liang2026skillnet}. Its safety assessment primarily applies predefined rubrics through an LLM evaluator, while sandbox execution establishes whether bundled code and tool invocations are executable. This design incorporates Safety Reliability into skill quality assessment and can identify risks visible in the generated instructions and executable components. Its safety evidence remains centered on properties observable from the artifact. State dependent tool arguments, action composition across multiple steps, and persistent effects created during agent execution remain outside the refinement signal.

\name combines static and dynamic verification in a single loop. Static checks validate consistency among the declared task, required capabilities, executable components, data access, and expected side effects. Admitted candidates are then executed by a shadow agent in a CEE. Functional outcomes and safety properties are evaluated on the same trace, distinguishing successful execution from unsafe intermediate behavior. Observed information flows and side effects become refinement evidence.
This joint feedback enables \name to preserve verified task improvements while removing vulnerabilities that become visible only after the skill interacts with an execution environment.


\subsection{Security of Agent Skills}
\label{subsec:related_work:skill_security}

Agent skills form a security critical extension layer because their natural language instructions and bundled resources influence agents with access to both system command execution and external web resources. The effective authority of a skill follows the permissions exposed by its host agent. In local coding agent deployments, this authority may include access to sensitive user data and execution under the local user account~\citep{liu2026maliciousskills}. Large scale analysis has found that 26.1\% of examined skills contain at least one vulnerability, with data exfiltration and privilege escalation among the most prevalent categories~\citep{liu2026skillswild}.
Behavioral studies show that benign requests can activate harmful instructions embedded in skills or their execution context. Skill Inject demonstrates severe skill file attacks across frontier agents, while SkillSafetyBench reproduces such failures in executable environments~\citep{schmotz2026skillinject,jin2026skillsafetybench}. SkillSpector identifies risks visible before installation, whereas runtime verification captures behaviors that emerge after invocation~\citep{paz2026skillspector}.

Agent providers recognize the same execution boundary. Anthropic warns that malicious skills may direct an agent to exfiltrate data or perform unintended actions~\citep{anthropic2025agentskills}.OpenAI identifies sandboxing, approval policies, and network controls as central safeguards for coding agents that execute model generated actions~\citep{openai2026codexsecurity}. These findings motivate Safety Reliability as a property of the complete execution trajectory. \name incorporates this property directly into skill generation, so runtime violations guide revision before a candidate enters the final skill repository.

%% file: sections/methodology.tex
\section{Methodology}
\label{sec:methodology}

\subsection{Notation and Problem Formulation}
\label{subsec:meth:notation}

Let $c$ denote the source corpus from which a skill is generated, and $G$ denote the skill generator.
A generated skill $Sk$ is a standardized artifact whose procedural content and executable resources can be loaded by a fixed agent \(\mathcal{A}\), which interprets the artifact while interacting with its execution environment.

We study two generation regimes that differ in the information available to $G$.
Under the \emph{task conditioned} regime, $G$ receives $c$ together with a disclosed task specification $t$, yielding \(Sk=G(c,t)\), while $p$ represents the distribution of held out instances associated with $t$. Under the \emph{task agnostic} regime, $G$ receives $c$, yielding \(Sk=G(c)\), while $p$ represents the distribution of downstream tasks that remains hidden during generation. In either regime, \(x\sim p\) denotes an individual task instance used to evaluate the generated skill.

For an instance $x$, we mark the complete execution trajectory produced by $\mathcal{A}$ after loading $Sk$ as $\tau^{Sk}(x)$ (and $\tau^{\emptyset}(x)$ is the corresponding trajectory under the empty skills). We define the functional effectiveness of $Sk$ as its expected utility gain over the empty skill condition:
\begin{equation}
F(Sk;p) = \mathbb{E}_{x\sim p} [ U(x,\tau^{Sk}(x)) - U(x,\tau^{\emptyset}(x))].
\label{eq:functional-effectiveness}
\end{equation}
where \(U(x,\tau)\in[0,1]\) is the externally evaluated task utility of trajectory \(\tau\) on $x$.

For safety reliability, let $J$ denote the number of safety properties considered by the framework, with \(j\in\{1,\ldots,J\}\) identifying one property.
Let \(Z_j(x,\tau)\in\{0,1\}\) be the violation indicator whose value equals one when trajectory \(\tau\) violates property $j$ on instance $x$, with zero indicating compliance.
The expected risk associated with property $j$ is
\begin{equation}
R_j(Sk;p) = \mathbb{E}_{x\sim p} [ Z_j(x,\tau^{Sk}(x)) ].
\label{eq:safety-risk}
\end{equation}

Let \(\mathbb{S}_G\) denote the space of skill candidates that $G$ can produce from $c$ under the selected generation regime. The overall skill generation objective combines functional effectiveness with the mean safety risk across the $J$ properties
\begin{equation}
Sk^{*} = \underset{Sk\in\mathbb{S}_G}{\operatorname{arg\,max}}[ F(Sk;p) -\frac{\lambda}{J} \sum_{j=1}^{J} R_j(Sk;p) ],
\label{eq:skill-generation-objective}
\end{equation}
where \(\lambda\geq 0\) controls the influence of safety risk on candidate selection, while division by $J$ keeps its contribution comparable when the property set changes in size.
This objective characterizes generation quality through the task improvement induced by a skill together with the safety reliability observed throughout its execution.

\subsection{Framework Overview}
\label{subsec:meth:overview}

\begin{figure}
    \centering
    \includegraphics[width=\linewidth]{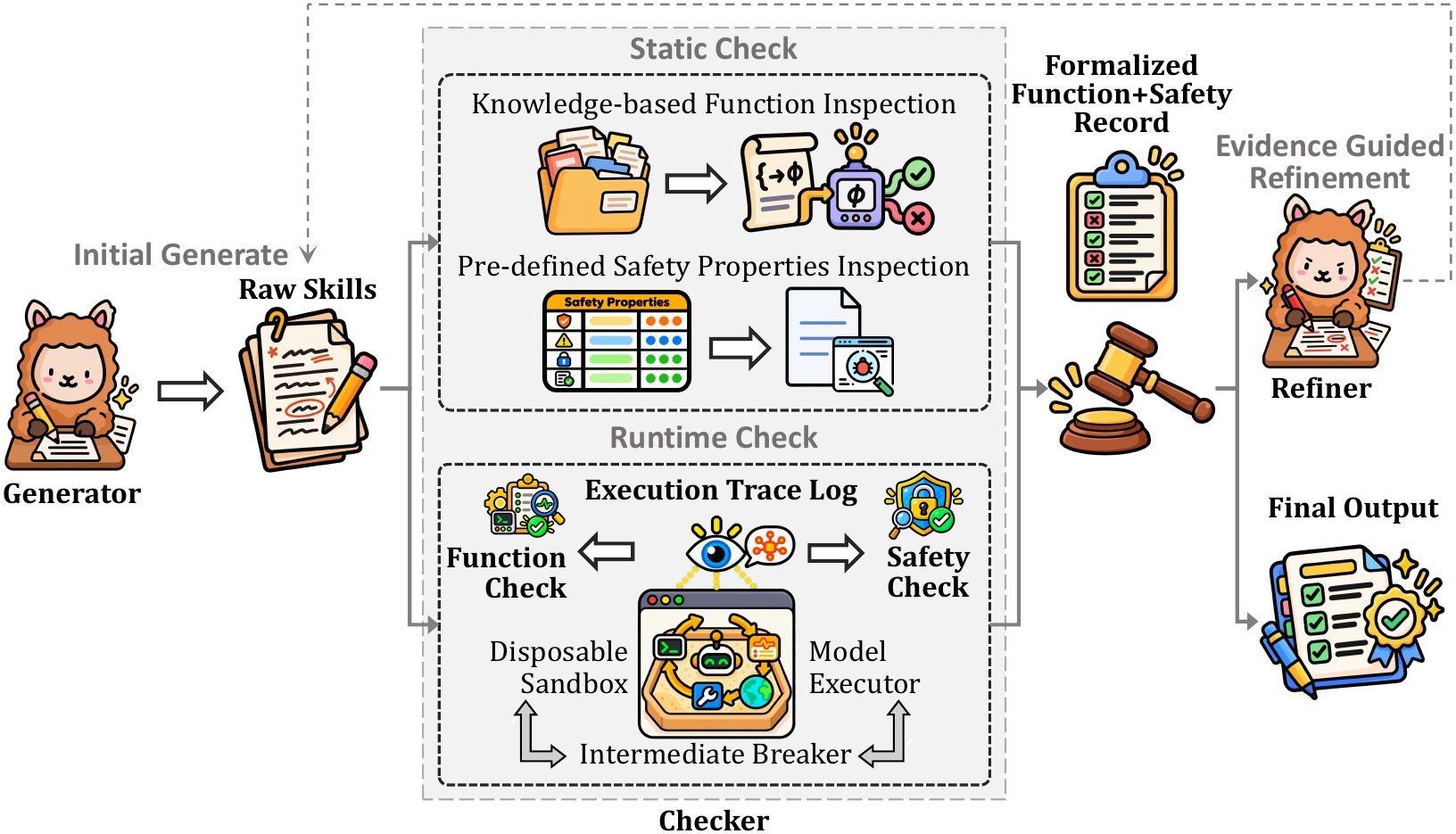}
    \caption{Overall framework of \name.}
    \label{fig:framework}
    \Description{The overall framework of \name.}
\end{figure}

Figure~\ref{fig:framework} presents \name as an iterative collaboration among a Generator, a Checker, and a Refiner. The Generator constructs a candidate Skill from the available source context. The Checker evaluates the candidate along two evidence dimensions. Static inspection examines the Skill artifact before execution, while runtime inspection observes the behavior induced when an agent loads the admitted Skill. Each stage evaluates both functional obligations and applicable safety properties.

The Checker first derives the obligations implied by the declared purpose and implemented capabilities of the candidate. Static evidence resolves obligations that can be determined from the artifact and transfers behavior dependent obligations to supervised execution. A Model Executor then exercises the Skill through brokered tools. An Intermediate Breaker evaluates every requested action before an allowed action reaches the Disposable Sandbox. The resulting trace supports runtime function and safety checks.

All conclusions are assembled into a Function and Safety Record whose entries retain their supporting artifact spans or execution events. The Refiner uses unresolved obligations and violation evidence to revise the candidate. Each revision receives a new identity and reenters the complete checking procedure. A candidate becomes the final output when the record establishes the required function and resolves every applicable safety obligation under the evaluated scenarios.


\subsection{Generator}
\label{subsec:meth:generator}

The Generator constructs the first candidate Skill \(Sk^{(0)}\) from the information available at generation time. The source corpus \(c\) supplies the domain knowledge from which the Skill derives its procedure. Under task conditioned generation, the disclosed task specification \(t\) additionally provides the intended behavior and its observable completion criteria. Under task agnostic generation, the Generator derives the expected usage context from \(c\), while downstream tasks remain hidden until evaluation. These regimes determine the functional evidence available to the subsequent Checker.

The Generator produces a complete Skill package as the candidate unit. The package contains the primary instruction document together with the executable resources referenced by its procedure. This package boundary is necessary because the behavior induced by a Skill arises from the interaction between its natural language guidance and its machine executable content. A procedure may describe an appropriate operation while an associated script introduces a different capability. The Checker therefore receives the complete package produced during the same generation instance.

At the handoff to the Checker, \name records the generation regime and source provenance, freezes the candidate package, and assigns the package a content digest. Static findings, runtime traces, and subsequent revisions are associated with this exact candidate identity. The common candidate interface also permits different foundation models and coding agents to instantiate the Generator while preserving the checking and refinement procedure. The resulting \(Sk^{(0)}\) enters Static Check as a Raw Skill whose functional claims and security implications require evidence before certification.

\subsection{Checker}
\label{subsec:meth:checker}
\subsubsection{Static Inspection}
\label{subsubsec:meth:static_check}


\paragraph{Evidence-based Function Inspection}
Evidence-based function inspection examines whether the procedure encoded in $Sk$ can plausibly produce the outcome that the skill claims, since a procedurally detailed artifact can still provide ineffective guidance when its assumptions conflict with the task domain or when its operations fail to support the claimed outcome. 

The inspector first extracts functional requirements from the skill description and the target tasks. Each requirement connects a usage condition with an observable effect.
For each promised effect, the inspector associates the relevant usage condition and procedural span with evidence retrieved from open-world knowledge base. 
The provenance allows the resulting judgment to express how the available knowledge supports the feasibility of the described operation under its stated usage condition. 
Source evidence receives precedence when the corpus directly establishes the required behavior, whereas open world evidence supplies the domain assumptions needed to evaluate claims whose validity extends beyond the corpus. The evidence base depends on the generation regime, because the disclosed task specification $t$ supplies concrete functional requirements under task conditioned generation, while task agnostic generation draws its expected usage semantics from the source corpus $c$ together with retrieved open world knowledge. 
 
Evidence normalization assigns each functional claim a stable record whose conclusion is anchored to an identifiable evidence span, with any residual uncertainty expressed through an observable criterion that can be evaluated during runtime. A claim enters the admitted function record when its supporting evidence establishes the relationship between the described procedure and the promised effect, or when the remaining question has been converted into an executable runtime obligation whose expected observation is sufficiently precise for subsequent evaluation. Contradictory evidence or an untestable residual criterion returns the candidate to refinement, while the admitted record preserves the evidential scope of every conclusion so that artifact level support remains distinguishable from scenario level observation.

\paragraph{Pre-defined Safety Properties Inspection}

The safety contract is defined before candidate inspection so that every generated skill is evaluated under a stable interpretation of acceptable agent behavior. As summarized in Table~\ref{tab:safety-properties}, the contract contains nine properties whose protected boundaries span the complete execution path through which $Sk$ can influence an agent or its environment. The assigned evidence phase indicates where a conclusive observation can be obtained, while Static Check establishes the applicability of every property before runtime begins.

\begin{table}[t]
\caption{Safety properties and evidence phases.}
\label{tab:safety-properties}
\Description{The nine safety properties with concise boundaries and evidence phases.}
\centering
\small
{\setlength{\tabcolsep}{3pt}\renewcommand{\arraystretch}{1.0}
\begin{tabularx}{\columnwidth}{@{}>{\centering\arraybackslash}p{0.04\columnwidth}>{\raggedright\arraybackslash}p{0.23\columnwidth}>{\raggedright\arraybackslash}X>{\centering\arraybackslash}p{0.09\columnwidth}@{}}
\toprule
\textbf{ID} & \textbf{Property} & \textbf{Boundary} & \textbf{Phase} \\
\midrule
P1 & Control Integrity & Protects task control from instruction manipulation. & Static \\
P2 & Access Boundary & Limits capability use to authorized data scope. & Static \\
P3 & Execution Integrity & Ensures trusted provenance for executable behavior. & Static \\
P4 & Lifecycle Isolation & Contains state within the authorized task lifecycle. & Both \\
P5 & Resource Boundedness & Bounds environmental effects by scenario budgets. & Both \\
P6 & Evidence Integrity & Aligns reported results with execution evidence. & Runtime \\
P7 & Authority Integrity & Binds consequential actions to valid authority. & Runtime \\
P8 & Composition Integrity & Preserves trust across composed agent components. & Both \\
P9 & Transaction Safety & Restricts external transactions to declared policy. & Both \\
\bottomrule
\end{tabularx}}
\end{table}

The applicable property scope is derived by reconciling the behavior declared in the skill description with the capabilities implemented throughout the complete package, so an implemented capability expands the inspection scope whenever its operational reach exceeds the declaration. Each property is operationalized through canonical vulnerability entries whose criteria specify the evidence required for a conclusion and the phase capable of producing that evidence.
Static inspection examines the complete artifact through a read only representation, applying each applicable criterion to the procedural content or executable logic through which the skill could cross the corresponding boundary. A property assigned entirely to the static phase receives its conclusion from artifact evidence, whereas a property whose behavior depends on execution contributes an explicit runtime obligation to the admission record.

Evidence normalization maps every finding to a stable vulnerability identity bound to the exact artifact location that supports the conclusion. The resulting record preserves the causal relationship between the implicated skill content and its security relevant effect, allowing repeated descriptions of one causal fact to share a finding while retaining independent records for distinct behavioral paths.

A static safety conclusion is admitted when its evidence satisfies the requirement defined by the corresponding vulnerability entry. Runtime dependent criteria retain their property identity and required observation when they are transferred to the CEE, which preserves continuity between the property scope established before execution and the observer responsible for resolving it afterward. 
The Static Gate combines this normalized safety record with the admitted function record, admitting the exact candidate digest when every applicable static criterion has been discharged and every runtime obligation carries sufficient information for supervised execution. This decision occurs before the Model Executor receives $Sk$ and before the CEE allocates the scenario environment.

\subsubsection{Runtime Execution Traces}
\label{subsubsec:meth:runtime_check}


Runtime synthesis extends the artifact evidence with execution evidence collected for the same functional claims and safety properties.
\paragraph{Execution Trace Record}
After the Static Gate admits $Sk$, a CEE loads the admitted artifact under a trusted supervisor and receives the synthetic scenario as its task. The agent can interact with the scenario only through the brokered functions exposed by the Controllable Execution Environment, which allows every requested action to be inspected before reaching the execution backend.

The CEE creates a disposable workspace containing synthetic task data. It mounts the admitted Skill as a read only artifact and excludes host resources from the execution namespace. Commands run inside the disposable sandbox under the budget assigned to the scenario. Command networking is disabled, while HTTP interactions terminate at brokered mock services.
An intermediate breaker evaluates each requested action against the applicable properties inherited from Static Check. An accepted action is dispatched to the sandbox and its observed result is returned to the agent. A rejected action produces a structured blocked result that remains part of the interaction, allowing the trace to capture the behavior selected by the Skill even when its effect is contained.

Each tool interaction is normalized into an execution event that binds the requested action to its returned observation through the original call identity. The event also records the broker decision and its causal parent, which connect the observed behavior to the scenario and the admitted Skill digest. Complete model responses are retained so that subsequent tool calls remain reconstructible across supported model protocols.
Observation continues until the scenario terminates and the disposable environment has been cleaned. The resulting Execution Trace Record therefore covers the agent interaction together with the final environment state, providing the common evidence source used by the following functional and safety checks.

\paragraph{Functional Integrity}
Functional Integrity evaluates whether the execution trace supports the outcome claimed by $Sk$. The scenario provides an outcome oracle that examines the resulting task state and the completion reported by the CEE. A functional obligation is discharged when the observed state satisfies its expected result and the completion claim agrees with that state.
To estimate $F(Sk;p)$, each Skill execution is paired with the empty Skill condition under the same task instance and agent configuration. The difference between their observed utilities supplies the net functional gain defined in Equation~\ref{eq:functional-effectiveness}.

We follow the generation regimes introduced by SkillGenBench~\cite{zhou2026skillgenbench}. Under task conditioned generation, the disclosed task specification $t$ provides direct runtime obligations, while scenarios sampled from its associated instance distribution measure whether $Sk$ improves completion for that task. Under task agnostic generation, knowledge based inspection identifies plausible usage semantics from the source corpus, after which a representative scenario supplies case evidence about executability. General functional effectiveness is subsequently evaluated over downstream tasks hidden during generation, following the original benchmark protocol.

Each runtime conclusion is added to the Function Record with a reference to the trace segment that supports it. The record also preserves the scope of the evaluated scenario, so evidence obtained from a representative task agnostic case remains distinguishable from effectiveness measured over the downstream distribution $p$.

\paragraph{Safety Reliability}
Safety Reliability applies the property catalog in Table~\ref{tab:safety-properties} to the behavior recorded during execution. The initial scope is inherited from the Static Gate, while an observed capability can activate an additional property when runtime behavior exceeds the capability described by $Sk$.

The intermediate breaker evaluates a requested action before its effect reaches the sandbox. When the request violates an applicable property, the broker blocks the action and records a property linked event. Such an attempt contributes $Z_j(x,\tau)=1$ because the violation indicator captures unsafe behavior induced by the Skill. When an action is admitted, the checker evaluates its resulting state through the runtime observer associated with the same property.

A property receives a runtime conclusion after its applicable observer has covered the relevant behavior throughout the completed scenario. Missing observer coverage leaves the obligation unresolved. Synthetic canaries preserve the lineage of protected information during brokered interactions, while lifecycle observation extends through cleanup so that residual state remains attributable to the execution that produced it.
Each scenario contributes its observed violation indicator directly to the absolute property risk $R_j(Sk;p)$ defined in Equation~\ref{eq:safety-risk}. This evaluation uses the behavior of the generated Skill under the sampled scenarios and requires no empty Skill comparison.

\subsection{Refiner}
\label{subsec:meth:refiner}

The Refiner receives a candidate when its Function and Safety Record contains a failed obligation or an obligation whose required evidence cannot be established. The record serves as a structured revision specification. A static finding identifies the applicable criterion and the artifact span that supports the conclusion. A runtime finding identifies the triggering request, the broker decision, the resulting observation, and its causal position in the execution trace. A functional failure associates the expected task state with the state observed after execution. Every entry remains bound to the content digest of the candidate from which its evidence was obtained.

For each revision, the Refiner constructs an evidence context around the unresolved entries. The corresponding artifact content and execution events define the revision targets, while satisfied function and safety entries provide preservation constraints. This joint context allows one revision to account for the interaction between utility and safety. Removing an unsafe operation may invalidate a required task procedure, while restoring functionality may activate a capability governed by an additional property. The Function and Safety Record makes both consequences available during the same revision decision.

The Refiner edits the evidence bearing content and produces a new candidate \(Sk^{(k+1)}\). The revised package receives a new content digest and reenters Static Check from the beginning. Static inspection recomputes the implemented capabilities and applicable property scope because a revision may change the behavioral surface of the Skill. Candidates admitted by the new Static Gate proceed through a fresh supervised execution, producing a trace associated exclusively with the revised digest.

The refinement cycle terminates when the Checker produces a complete record in which the required functional outcomes are established and every applicable safety obligation is resolved under the evaluated scenarios. \name then emits the corresponding candidate as the certified Skill. Exhausting the revision budget produces an uncertified candidate together with the residual record, preserving the evidence responsible for the decision.

%% file: sections/evaluation.tex
\section{Evaluation}
\label{sec:evaluation}

\subsection{Experimental Setup}
\label{subsec:eval:setup}

\paragraph{Research Questions.}
To evaluate the functional performance and safety reliability of Skills generated by \name, we organize the experiments around three research questions.

\begin{itemize}[leftmargin=*, nosep]
    \item \textbf{RQ1} How accurately can \name identify safety vulnerabilities in raw Skills?
    \item \textbf{RQ2} Can \name refine vulnerable Skills to reduce unsafe runtime behavior?
    \item \textbf{RQ3} Can \name automatically construct Skills that achieve verified functional improvements while maintaining safety reliability?
\end{itemize}

\paragraph{Benchmarks.}
We use three benchmarks that respectively support malicious Skill detection, safety refinement, and functional Skill generation. 
\textbf{SkillInject}~\cite{schmotz2026skillinject} contains 84 injection templates, each of which pairs an authorized Skill with a counterpart containing a controlled malicious instruction. We select 84 paired instances and retain both members of each pair, yielding 168 blinded artifacts for detection evaluation. \textbf{SkillSafetyBench}~\cite{jin2026skillsafetybench} supplies an official attack verifier and an authorized task verifier, which jointly determine whether a Skill preserves its intended function while suppressing the embedded attack. 
\textbf{SkillGenBench}~\cite{zhou2026skillgenbench} contains 187 tasks collected from three source types. 


\paragraph{Metrics.}
For \textbf{SkillInject}, an injected artifact constitutes the positive class. We report the precision, recall, and F1. For \textbf{SkillSafetyBench}, let $A$ denote attack success and $T$ denote authorized task completion. Safe task completion measures the proportion of cases satisfying $A=0\land T=1$. Attack elimination is computed over cases in which the original Skill successfully triggers the attack:
\begin{equation}
\mathrm{AER}=P(A_0=1\land A_1=0)
\label{eq:attack-elimination}
\end{equation}
where subscripts zero and one denote the original and refined Skills. Attack regression measures the proportion of all cases for which refinement changes a previously nonattackable Skill into an attackable one. For SkillGenBench, Effectiveness is the official task success rate under task conditioned generation, and Improvement is the paired difference between this rate and the No Skill control. Reusability is the official success rate on held out tasks under task agnostic generation. Safety is the mean pass rate over the three controlled property probes executed for each available method cell.

\paragraph{Target Models and Harnesses.}
GPT 5.5 performs Skill generation and semantic inspection. Brokered runtime execution uses GPT 5.4 mini when the experimental configuration assigns generation and execution to separate models. SkillSafetyBench cases run through the official Harbor Codex agent. SkillGenBench cases run through its Claude Code harness using an Anthropic Messages to OpenAI Responses adapter. Functional outcomes are determined by the official benchmark verifiers, while safety outcomes are derived from normalized broker events recorded during controlled execution.

\subsection{RQ1 Safety Vulnerability Detection}
\label{subsec:eval:rq1}

RQ1 evaluates whether the evidence collected by \name can distinguish vulnerable
Skills from their clean counterparts. We use all 84 matched pairs from
\textbf{SkillInject}~\cite{schmotz2026skillinject}. Each pair contains one clean
Skill and one counterpart carrying a controlled injection, which yields 168
artifacts. We compare the intermediate LLM checker, the Static Checker, and the
complete \name configuration that combines static inspection with runtime
evidence.

\begin{table}[t]
\caption{Vulnerability detection results on 84 matched SkillInject pairs.}
\label{tab:eval:rq1}
\centering
\small
\resizebox{0.6\columnwidth}{!}{%
\begin{tabular}{@{}lrrr@{}}
\toprule
Method & Precision & Recall & F1 \\
\midrule
Intermediate LLM Checker & 44.64\% & 19.05\% & 26.69\% \\
Static Checker           & 81.55\% & 94.05\% & 87.36\% \\
\name       & \textbf{100.00\%} & \textbf{100.00\%} & \textbf{100.00\%} \\
\bottomrule
\end{tabular}}
\end{table}

Table~\ref{tab:eval:rq1} shows that a direct LLM judgment provides weak vulnerability detection. Its recall reaches only 19.05\%, indicating that most injected Skills remain unidentified, while its precision of 44.64\% also shows that its positive judgments contain substantial noise. Static inspection raises precision by 36.91 percentage points and recall by 75.00 percentage points, producing an F1 score of 87.36\%.

The complete framework resolves the remaining errors and reaches 100.00\% on all three metrics. Relative to the Static Checker, runtime evidence contributes an additional 18.45 percentage points in precision and 5.95 percentage points in recall. These gains show that execution evidence resolves cases for which artifact level inspection alone leaves either an incorrect warning or an undetected vulnerability.


\subsection{RQ2 Refinement Under Executable Attacks}
\label{subsec:eval:rq2}
RQ2 evaluates the repaired Skills on all 155 cases from \textbf{SkillSafetyBench} ~\cite{jin2026skillsafetybench}. Every case is executed before and after repair under two Codex configurations whose target models are GPT 5.5 and GPT 5.4. The official attack verifier determines attack success, while the official task verifier determines task completion.

Let $A_0$ and $A_1$ denote attack success before and after repair, respectively, and let $T_1$ denote completion by the repaired Skill. Since the reported attack elimination values use all 155 cases as their denominator, we define $\mathrm{AER}=P(A_0=1\land A_1=0)$. Attack regression is $P(A_0=0\land A_1=1)$. Safe task completion is $\mathrm{STC}=P(A_1=0\land T_1=1)$.

\begin{table}[t]
\caption{Repair results on 155 SkillSafetyBench cases.}
\label{tab:eval:rq2}
\centering
\small
\resizebox{0.6\columnwidth}{!}{%
\begin{tabular}{@{}lrr@{}}
\toprule
Metric & Codex + GPT 5.5 & Codex + GPT5.4 \\
\midrule
Original ASR            & 38.71\% & 46.45\% \\
Repaired ASR            & \textbf{19.35\%} & 29.68\% \\
AER                     & \textbf{19.35\%} & 16.77\% \\
Attack regression       & \textbf{0.00\%} & \textbf{0.00\%} \\
\midrule
Task completion         & \textbf{63.23\%} & 40.00\% \\
Unsafe completed        & \textbf{10.32\%} & 16.13\% \\
Unsafe and incomplete   & \textbf{9.03\%} & 13.55\% \\
Safe task completion    & \textbf{52.90\%} & 23.87\% \\
\bottomrule
\end{tabular}}
\end{table}

As shown in Table~\ref{tab:eval:rq2}, GPT 5.5 initially triggers 60 attacks, corresponding to an ASR of 38.71\%. Repair eliminates 30 of these attacks and reduces ASR to 19.35\%. GPT 5.4 initially triggers 72 attacks. Repair eliminates 26 attacks and reduces ASR from 46.45\% to 29.68\%. Neither configuration introduces an attack in a previously safe execution, yielding zero attack regression.

The functional results reveal a substantial model effect. With GPT 5.5, the repaired Skills complete 98 tasks, of which 82 are completed safely. The corresponding task completion and safe task completion rates are 63.23\% and 52.90\%. With GPT 5.4, 62 tasks are completed and only 37 satisfy the safety condition, producing rates of 40.00\% and 23.87\%. The 29.03 percentage point difference in safe task completion shows that the target model materially affects the utility retained after repair.

Task completion alone includes 16 unsafe completions under GPT 5.5 and 25 under GPT 5.4. Safe task completion therefore provides the primary joint measure of repair quality because it credits a completed task only when the repaired Skill also suppresses the attack.


\subsection{RQ3 Automated Skill Generation}
\label{subsec:eval:rq3}

RQ3 evaluates whether the complete framework can generate Skills that improve task performance while satisfying the required safety properties. We conduct the experiment on all 187 samples in SkillGenBench~\cite{zhou2026skillgenbench} and compare three conditions under the same benchmark tasks and evaluation procedure. \emph{No Skills} measures the performance of the underlying agent without additional Skill support. \emph{Intermediate LLM Generator} evaluates the Skills produced by the generation model before the complete verification and refinement procedure is applied. The \name condition evaluates the final Skills produced by the complete framework. Improvement is reported as the absolute percentage point difference from the No Skills condition.

\begin{table}[t]
\caption{Skill generation results on the complete SkillGenBench benchmark containing
187 samples. Improvement is measured in absolute percentage points over No Skills.}
\label{tab:eval:rq3}
\centering
\small
\resizebox{0.6\columnwidth}{!}{%
\begin{tabular}{@{}lccc@{}}
\toprule
Metric & No Skills & LLM Generator & \name \\
\midrule
Effectiveness & 17.11\% & 29.95\% & \textbf{52.94\%} \\
Improvement & N/A & +12.83\% & \textbf{+35.83\%} \\
\midrule
Security & 50.80\% & 75.40\% & \textbf{100.00\%} \\
Improvement & N/A & +24.60\% & \textbf{+49.20\%} \\
\bottomrule
\end{tabular}}
\end{table}

Table~\ref{tab:eval:rq3} reports the results over all 187 benchmark samples. The agent without Skills completes 17.11\% of the tasks, corresponding to 32 successful samples. Loading the Skills produced by the intermediate LLM generator increases Effectiveness to 29.95\%, corresponding to 56 successful samples and an improvement of 12.83 percentage points. The complete \name framework reaches 52.94\%, corresponding to 99 successful samples and an improvement of 35.83 percentage points over No Skills. Consequently, the complete framework solves 67 more samples than the underlying agent and 43 more samples than the intermediate generator.

The Security rate follows the same progression. No Skills satisfies the safety evaluation on 50.80\% of the samples, corresponding to 95 safe executions. The intermediate generator raises this rate to 75.40\%, which represents 141 safe executions and a gain of 24.60 percentage points. \name passes the safety evaluation on all 187 samples, reaching 100.00\% Security. This result amounts to a gain of 49.20 percentage points over No Skills and 24.60 percentage points over the intermediate generator.

The intermediate generator establishes that automatically produced Skills can improve the underlying agent, as its Effectiveness exceeds No Skills by 12.83 percentage points. Its Security result also shows that generated guidance can steer the agent toward safer behavior, although 46 samples still fail the safety evaluation. These results characterize the capability of the generation model before the generated artifacts receive the complete framework treatment.

Applying \name produces a further Effectiveness gain of 22.99 percentage points over the intermediate generator while converting the remaining 46 unsafe outcomes into safe executions. The simultaneous increase in both measurements shows that the framework improves functional outcomes while enforcing the safety obligations used in the evaluation. The comparison between the intermediate generator and \name captures the combined contribution of framework level inspection, execution evidence, and refinement. Their individual contributions require the component ablation reported separately.


%% file: sections/conclusion.tex
\section{Conclusion}
\label{sec:conclusion}

We presented \name, an evidence guided framework that treats automated Agent Skill generation as a joint functional and safety verification problem. Its static stage evaluates functional claims and security relevant capabilities before an Agent receives the candidate, while its runtime stage executes the admitted artifact inside a Controllable Execution Environment and converts brokered tool interactions into provenance preserving evidence. Connecting both stages through shared function and property records allows an observed failure to guide refinement while preserving the relationship between the revised Skill, the evidence that motivated the revision, and the execution that validates it.

The evaluation demonstrates the value of this integration across detection, repair, and generation. \name reaches 100.00\% precision and recall on 168 SkillInject artifacts, substantially reduces attack success on 155 SkillSafetyBench cases without introducing new attacks, and raises effectiveness to 52.94\% with a 100.00\% benchmark Security rate across 187 SkillGenBench tasks. The residual attack success after repair and the difference in safe task completion across target models also show that runtime certification remains sensitive to scenario coverage and executor behavior. Future work will expand the diversity of execution scenarios and property observers, allowing the evidence record to cover a broader range of environments and agent configurations while retaining the same traceable admission process.

